\documentclass{article}
\usepackage{textcomp}
\usepackage{spconf}

\usepackage{graphicx}

\usepackage{amsmath}
\usepackage{amssymb}
\usepackage{gensymb}

\usepackage{cite}  
\usepackage{tabularx}
\usepackage{booktabs}
\usepackage{multirow}
\usepackage{makecell}
\usepackage[acronym]{glossaries}

\usepackage{caption}

\usepackage[breaklinks=true,bookmarks=false]{hyperref}

\newacronym{cnn}{CNN}{Convolutional Neural Network}
\newacronym{gcn}{GCN}{Graph Convolutional Network}
\newacronym{gei}{GEI}{Gait Energy Image}
\newacronym{cvae}{CVAE}{Conditional Variational Autoencoder}
\newacronym{sota}{SOTA}{state-of-the-art}

\newcommand{\nparagraph}[1]{\noindent\textbf{#1.  }}

\title{GaitGraph: Graph Convolutional Network for \\Skeleton-Based Gait Recognition}
\name{
Torben Teepe
\quad Ali Khan
\quad Johannes Gilg
\quad Fabian Herzog
\quad Stefan H\"ormann
\quad Gerhard Rigoll
\thanks{\copyright 2021 IEEE. Personal use of this material is permitted. Permission from IEEE must be obtained for all other uses, in any current or future media, including reprinting/republishing this material for advertising or promotional purposes, creating new collective works, for resale or redistribution to servers or lists, or reuse of any copyrighted component of this work in other works.}
}
\address{Technical University of Munich}

\begin{document}
%
\maketitle
\begin{abstract}
Gait recognition is a promising video-based biometric for identifying individual walking patterns from a long distance.
At present, most gait recognition methods use silhouette images to represent a person in each frame. However, silhouette images can lose fine-grained spatial information, and most papers do not regard how to obtain these silhouettes in complex scenes. Furthermore, silhouette images contain not only gait features but also other visual clues that can be recognized. Hence these approaches can not be considered as strict gait recognition.

We leverage recent advances in human pose estimation to estimate robust skeleton poses directly from RGB images to bring back model-based gait recognition with a cleaner representation of gait. Thus, we propose \textit{GaitGraph} that combines skeleton poses with \gls{gcn} to obtain a modern model-based approach for gait recognition. The main advantages are a cleaner, more elegant extraction of the gait features and the ability to incorporate powerful spatio-temporal modeling using \gls{gcn}.
Experiments on the popular CASIA-B gait dataset show that our method archives state-of-the-art performance in model-based gait recognition.

The code and models are publicly available\footnote{\href{https://github.com/tteepe/GaitGraph}{\tt github.com/tteepe/GaitGraph}}.

\end{abstract}
%

\section{Introduction}
\label{sec:intro}
Compared to other unique biometrics like face, fingerprint, and iris, gait is remarkable in the recognition from a great distance and without the cooperation or intrusion to the subject. Hence, it opens up enormous potential for applications such as social security, access control, and forensic identification.

However, gait can be sensitive to surface type, clothing, carried items, and clutter or occlusions in the scene.
These represent the challenges of tackling the gait identification task to learn unique and invariant features from the human gait.

Most current approaches \cite{song2019gaitnet, chao2019gaitset, fan2020gaitpart} use silhouettes extracted from a video sequence to represent the gait. These approaches apply the following steps: silhouette extraction, feature learning, and similarity comparison. The silhouette extraction is mostly done using background subtraction \cite{wang2003silhouette}. While background subtraction is easy to apply in a lab setting, it becomes cumbersome in a cluttered and rapidly-changing real-world scenario. Most applications \cite{chao2019gaitset, fan2020gaitpart, wu2016comprehensive} do not consider the complexity of the background subtraction task. Other approaches go up to the extent of training a separate \gls{cnn} for this task \cite{song2019gaitnet}.

\begin{figure}[t!]
  \begin{center}
  \includegraphics[width=0.785\linewidth]{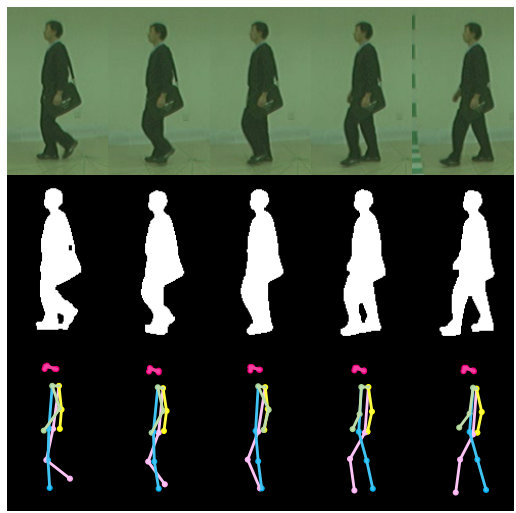}
  \end{center}
  \caption{Comparison of different gait representations of a subject in the \mbox{CASIA-B} gait dataset at different timesteps. Each row depicts the same frames as RGB image, silhouette image, and 2D skeleton pose, respectively, from top-to-bottom.}
  \label{fig:gaitseq}
\end{figure}

With robust human pose estimators emerging, other approaches \cite{liao2017pose, liao2020model} started new model-based methods for gait. Current pose estimation algorithms are very robust against occlusion, cluttered and changing backgrounds, carried items, and clothing. Compared to silhouette images, multiple poses can be extracted from an image simultaneously, even if they overlap \cite{cheng2020bottom}. Pose estimation in 2D and 3D is an active area of research, and our approach will profit from further improvements.
Furthermore, a skeleton sequence is a \textit{cleaner} representation of the gait since silhouette images also capture visual information of the person like physique or hairstyle. Hence, silhouette-based approaches recognize gait features and other appearance clues, which make these approaches more comparable to person re-identification methods.

This paper proposes GaitGraph, a novel approach where we apply a \gls{gcn} on a graph of human skeleton poses.
Inspired by the success of \glspl{gcn} in skeleton-based action recognition \cite{yan2018spatial, song2020stronger}, we adapted the methods to the gait recognition task.
The pose estimation replaces the silhouette extraction from previous approaches. 
The skeleton-based representation brings back \textit{real} gait recognition while also using less sensitive personal data. 

Our contributions can be summarized as follows:

(1) We use a modern interpretation of model-based gait recognition, exploiting robust human pose estimation and powerful temporal and spatial modeling of \glspl{gcn}.

(2) Our empirical experiments show \gls{sota} results compared to the current model-based approaches and even competitive results compared to appearance-based methods.

\section{Related Work}
Current works in gait recognition can be grouped by their spatial feature extraction and their temporal modeling.
 
For the spatial feature extraction, there are two categories: appearance-based and model-based approaches.
Appearance-based methods relied on a binary human silhouette image extracted from the original image \cite{wang2003silhouette}. The extraction is usually obtained by background subtraction for static scenes but becomes more complicated for dynamic and changing settings \cite{song2019gaitnet}. While most approaches \cite{chao2019gaitset, wu2016comprehensive, song2019gaitnet} use the whole shape as input, recent methods \cite{fan2020gaitpart} focus on specific body parts.
Model-based approaches consider the underlying physical structure of the body \cite{bouchrika2007model, liao2017pose, liao2020model}. The features extracted from the model data are mostly handcrafted and contain velocity, angles, etc. While model-based approaches used to be computationally expensive, the advances in pose estimation have now made them an interesting possibility.

The temporal modeling can be divided into single-image, sequence-based, and set-based approaches. Early approaches proposed to encode a gait cycle into a single image, i.e., \gls{gei} \cite{han2005individual}. These representations are easy to compute but lose most of the temporal information. Sequence-based approaches focus on each input separately. For modeling the temporal information 3D-CNNs \cite{wolf2016multi, liao2017pose} or LSTMs \cite{sokolova2018pose, liao2020model} are used. These approaches can comprehend more spatial information and gather more temporal information but require higher computational costs. The set-based approach \cite{chao2019gaitset} with shuffled inputs models no temporal information, thus has less computational complexity. \textit{GaitPart} \cite{fan2020gaitpart} introduces a novel temporal module, that focuses on capturing short-range temporal features.

In recent years, some approaches \cite{liao2020model, an2018improving} extract skeleton features with a pose estimator for gait recognition. These approaches use traditional \glspl{cnn} combined with LSTMs \cite{an2018improving} or handcrafted temporal features \cite{liao2020model}.

\begin{figure*}[ht]
  \begin{center}
  \includegraphics[width=0.99\linewidth]{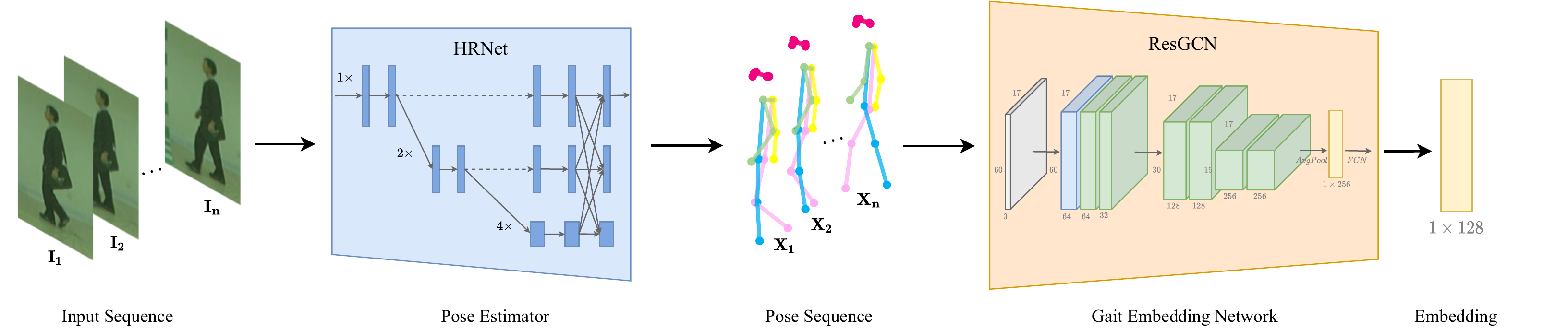}
  \end{center}
  \caption{\textbf{Overview of the Pipeline.} Starting with a sequence of images, for each image a pose is estimated. The sequence of poses is then feed through the ResGCN yielding the feature embedding.}
  \label{fig:pipeline}
\end{figure*}

\section{Skeleton-Based Gait Recognition}
In this section, we describe our method for learning discriminative information from a sequence of human poses. The overall pipeline is illustrated in Fig.~\ref{fig:pipeline}.

\subsection{Preliminaries}

\nparagraph{Notation} A human skeleton graph is denoted as $\mathcal{G} = (\mathcal{V}, \mathcal{E})$, where $\mathcal{V} = \{v_1, \hdots, v_N\}$ is the set of $N$ nodes representing joints, and $\mathcal{E}$ is the set of edges representing bones captured by an adjacency matrix $\mathbf{A} \in \mathbb{R}^{N \times N}$ with $\mathbf{A}_{i,j} = 1$ if an edge connects from $v_i$ to $v_j$ and $\mathbf{A}_{i,j} = 0$ otherwise.
$\mathbf{A}$ is symmetric since $\mathcal{G}$ is undirected.

Gait as a \textit{sequence of graphs} has a node feature set $\mathcal{X} = \{\boldsymbol{x}_{t,n} \in \mathbb{R}^{C} \mid t,n \in \mathbb{Z}, 1\leq t\leq T, 1\leq n \leq N\}$ represented as a feature tensor $\mathbf{X} \in \mathbb{R}^{T \times N \times C}$, where $\mathbf{x}_{t,n} = \mathbf{X}_{t,n,:}$ is the $C$ dimensional feature vector for node $v_n$ at time $t$ over a total of $T$ frames.

Thus, the input gait can be described by $\mathbf{A}$ structurally and by $\mathbf{X}$ feature-wise, with $\mathbf{X}_t \in \mathbb{R}^{N\times C}$ being a pose at time $t$. The pose feature $\mathbf{X}$ in the $C$ dimension is a tuple of 2D coordinate, and it's confidence. The $N$ dimension is the number of joints. A learnable weight matrix at layer $l$ of a network is denoted as $\Theta^{(l)} \in \mathbb{R}^{C_l \times C_{l+1}}$.\\
\\
\nparagraph{Graph Convolutions}
On skeleton inputs, defined by features $\mathbf{X}$ and graph structure $\mathbf{A}$, the layer-wise update rule of graph convolutions can be applied to features at time $t$ as:
\begin{equation} \label{eq:gcn}
    \mathbf{X}_t^{(l+1)} = \sigma\left(
        \tilde{\mathbf{D}}^{-\frac{1}{2}}
        \tilde{\mathbf{A}}
        \tilde{\mathbf{D}}^{-\frac{1}{2}}
        \mathbf{X}_t^{(l)}
        \Theta^{(l)}
    \right),
\end{equation}
where $\tilde{\mathbf{A}} = \mathbf{A + I}$ is the skeleton graph with added \textit{self-loops} to keep identity features, $\tilde{\mathbf{D}}$ is the diagonal degree matrix of $\tilde{\mathbf{A}}$, and $\sigma(\cdot)$ is an activation function.
The term
$
\tilde{\mathbf{D}}^{-\frac{1}{2}}
\tilde{\mathbf{A}}
\tilde{\mathbf{D}}^{-\frac{1}{2}}
\mathbf{X}_t^{(l)}
$
can be intuitively interpreted as an approximate spatial \textit{mean} feature aggregation from the messages passed by the direct neighbors.

\subsection{Human Pose Extraction}
For the feature extraction from the raw input images we estimate the human pose in each frame.
The pose estimation or simply a keypoint detection aims to detect the locations of $N$ keypoints (e.g., shoulder, hip, knee, etc.) from an image $\mathbf{I} \in \mathbb{R}^{W \times H \times 3}$. The \gls{sota} method \cite{cheng2020bottom} solve this problem by estimating $N$ heatmaps $\left\{\mathbf{H}_1, \mathbf{H}_2, \ldots, \mathbf{H}_N \right\}$ of size $W' \times H'$, where the heatmap $\mathbf{H}_n$ indicates the location of the $n$-th keypoint. The location of the maximum of these heatmaps $\mathbf{H}_n$ yields the location of the keypoint $v_n$ that define the edges $\mathcal{V}$.

In our approach we use \textit{HRNet} \cite{sun2019deep}\footnote{\href{https://github.com/HRNet/HRNet-Human-Pose-Estimation}{\tt github.com/HRNet/HRNet-Human-Pose-Estimation}} as a 2D human pose estimator. We use the provided network, which is pre-trained  on the COCO dataset \cite{lin2014microsoft}. The COCO dataset pose annotations consist of 17 keypoints. There is no provided set of bones or edges $\mathcal{E}$ but we use a commonly used configuration as shown in the last row of Fig.~\ref{fig:gaitseq}. 

\subsection{Network and Implementation Details}
The network's main architecture follows the design proposed as the \textit{ResGCN} in \cite{song2020stronger} with adaptions to our use case. The network is composed of ResGCN blocks. The block consists of a Graph Convolution followed by a \textit{classic} 2D Convolution in the temporal domain and a residual connection with an optional bottleneck structure. The network is then composed of multiple ResGCN blocks in sequence (see Tab.~\ref{tab:net} for detailed configuration), followed by an average pooling and a fully connected layer that is yielding the feature vector. As the loss function, we use supervised contrastive (\textit{SupCon}) loss \cite{khosla2020supervised}.

\nparagraph{Augmentation} For augmentation on the skeleton graph, we use multiple unique augmentation techniques. First, we flip the order of the sequence, which can be interpreted as the person walking backward. Secondly, we mirror the skeleton graph along a vertical axis through the graph's center of gravity. This augmentation causes the person to walk in the opposite direction. Furthermore, we add small Gaussian noise to each joint and the same joint in the sequence to make our network more robust to the pose estimation's inaccuracies.

\nparagraph{Testing} At testing, the distance between gallery and probe is defined as the Euclidean distance of the corresponding feature vectors. Besides, we feed the original and a flipped order sequence to the network and take the average of two feature vectors.

\begin{table}
 \caption{Overview of the \textit{ResGCN-N39-R8} network architecture for a pose with 17 joints and sequence length of 60.}
 \label{tab:net}

\fontsize{9}{11}\selectfont
 \centering
 \begin{tabularx}{.71\linewidth}{l  l  c }
 \toprule
  Block & Module & Output Dimensions \\
 \midrule 
  Block 0 & BatchNorm & $60 \times 17 \times 3$ \\
  \hline
  \multirow{3}{*}{Block 1}
   & Basic & $60 \times 17 \times64$\\
   & Bottleneck &  $60 \times 17 \times 64$\\
   & Bottleneck &  $60 \times 17 \times 32$\\
  \hline
  \multirow{4}{*}{Block 2}
   & Bottleneck &  $30 \times 17 \times 128$\\
   & Bottleneck &  $30 \times 17 \times 128$\\
   & Bottleneck &  $15 \times 17 \times 256$\\
   & Bottleneck &  $15 \times 17 \times 256$\\
   \hline
   \multirow{2}{*}{Block 3} & AvgPool2D & $1 \times 256$ \\
   & FCN & $1 \times 128$ \\
 \bottomrule
\end{tabularx}
\end{table}
\begin{table*}[tbp]
 \caption{Averaged Rank-1 accuracies in percent on CASIA-B per probe angle excluding identical-view cases compared with other model-based methods.}
 \label{tab:casia-b-model}
\centering
\fontsize{9}{11}\selectfont
\begin{tabularx}{.88\textwidth}{l|l|c c c c c c c c c c c|c}
     \toprule
     \multicolumn{2}{l|}{Gallery NM\#1-4}&\multicolumn{11}{c|}{0\degree-180\degree}&\multirow{2}{*}{mean}\\[0.3mm]

     \multicolumn{2}{l|}{Probe}& 0\degree&18\degree&36\degree&54\degree&72\degree&90\degree&108\degree&126\degree&144\degree&162\degree&180\degree\\
     
     \hline
     
     \multirow{2}{*}{NM\#5-6}
    
     &PoseGait \cite{liao2020model} & 55.3 & 69.6 & 73.9 & 75.0 & 68.0 & 68.2 & 71.1 & 72.9 & 76.1 & 70.4 & 55.4 & 68.7\\
     &\textbf{GaitGraph} & 85.3 &  88.5 &  91.0 &  92.5 &  87.2 &  86.5 &  88.4 &  89.2 &  87.9 &  85.9 &  81.9 &  87.7 \\
     
     \hline
     
     \multirow{2}{*}{BG\#1-2}
    
     &PoseGait \cite{liao2020model} & 35.3 & 47.2 & 52.4 & 46.9 & 45.5 & 43.9 & 46.1 & 48.1 & 49.4 & 43.6 & 31.1 & 44.5\\
     &\textbf{GaitGraph}              & 75.8 &  76.7 &  75.9 &  76.1 &  71.4 &  73.9 &  78.0 &  74.7 &  75.4 &  75.4 &  69.2 &  74.8 \\

     \hline
     
     \multirow{2}{*}{CL\#1-2}

     &PoseGait \cite{liao2020model} & 24.3 & 29.7 & 41.3 & 38.8 & 38.2 & 38.5 & 41.6 & 44.9 & 42.2 & 33.4 & 22.5 & 36.0 \\
     &\textbf{GaitGraph}            & 69.6 &  66.1 &  68.8 &  67.2 &  64.5 &  62.0 &  69.5 &  65.6 &  65.7 &  66.1 &  64.3 &  66.3 \\
     \bottomrule
\end{tabularx}
\end{table*}

\begin{table}
 \caption{Averaged Rank-1 accuracies in percent on CASIA-B comparison with both appearance-based and model-based methods.}
 \label{tab:casia-b-all}
\fontsize{9}{11}\selectfont
 \centering
 
 \begin{tabularx}{0.8\linewidth}{r | l | c c c }
 \toprule
  & & \multicolumn{3}{c}{Probe}\\
   Type &Method &  NM & BG & CL\\
   \hline
   \multirow{3}{*}{\makecell[r]{appearance\\-based}}&GaitNet \cite{song2019gaitnet} & 91.6 & 85.7 & 58.9\\
   &GaitSet \cite{chao2019gaitset} & 95.0 & 87.2 & 70.4\\
   &GaitPart \cite{fan2020gaitpart}& \textbf{96.2} & \textbf{91.5} & \textbf{78.7}\\
   \hline
   \multirow{2}{*}{\makecell[r]{model\\-based}}&PoseGait \cite{liao2020model} & 68.7 & 44.5 & 36.0\\
   &\textbf{GaitGraph}            & \textbf{87.7} & \textbf{74.8} & \textbf{66.3}\\
 \bottomrule
\end{tabularx}
\end{table}

\section{Experiments}
In this part, we compare GaitGraph to other \gls{sota} methods in public gait dataset CASIA-B \cite{yu2006framework}. We compare the performance on multiple views and multiple walking conditions with model-based and appearance-based methods and conduct ablation studies to evaluate our temporal and spatial modeling.

\subsection{Dataset and Training Details}
Most available gait datasets do not provide RGB images since they are tailored to gait methods that rely on silhouettes or \glspl{gei}.
Therefore we cannot evaluate on the largest public gait dataset \mbox{OU-MVLP} \cite{takemura2018multi}, the evaluation on the commonly used dataset CASIA-B provides a comparison with other methods.

\textbf{CASIA-B} \cite{yu2006framework} is a widely used gait dataset and composed of 124 subjects. For each of the 124 subjects the dataset contains 11 views (0\degree, 18\degree, \dots, 180\degree) and 3 waking conditions. The walking conditions are normal (NM) (6 sequences per subject), walking with a bag (BG) (2 sequences per subject), and wearing a coat or a jacket (CL) (2 sequences per subject). Summed up, each subject contains $11\times(6+2+2)=110$ sequences.

Since there is no official partition of training and test set, there are various experiment protocols \cite{zhang2019gait}. For a fair comparison, this paper follows the popular protocol by \cite{wu2016comprehensive}. Furthermore, we use the commonly called large-sample training (LT) partition. In LT, the first 74 subjects comprise the training set, whereas the remaining 50 subjects form the test set. In the test sets of all three settings, the first four sequences of the NM condition (NM \#1-4) are kept in the gallery, and the remaining six sequences are divided into three probe subsets, i.e., NM subsets containing NM \#5-6, BG subsets containing BG \#1-2 and CL subsets containing CL \#1-2.\\

\nparagraph{Training Details}
The pose sequence is partitioned as a graph using the spatial configuration as mentioned in \cite{yan2018spatial} with a sequence length $T=60$ frames. Adam optimizer is used with a \textit{1-cycle} learning rate \cite{smith2019super} and a weight decay penalty of 1e-5. For the first cycle, the maximum learning rate is set to 0.01 for 300 epochs, and for the second cycle, the maximum learning rate is 1e-5 for 100 epochs. The loss function's temperature is set to 0.01, and the batch size is 128.

\subsection{Comparison with State-of-the-Art Methods}
Tab~\ref{tab:casia-b-model} shows the comparison of GaitGraph to \textit{PoseGait} \cite{liao2017pose}, which represents the sole pose-based approach to gait recognition utilizing handcrafted pose features. Our approach indicates significant improvements throughout all cross-views and walking conditions. With both approaches using a similar performing pose extractor, this proves the superiority of our \gls{gcn} architecture as a feature extractor. 

The currently best performing models use appearance-based features. In Tab~\ref{tab:casia-b-all}, we compare the appearance-based and model-based methods with our approach. The first three methods all use explicitly \cite{chao2019gaitset, fan2020gaitpart} or implicitly \cite{chao2019gaitset} silhouette images as their feature representation. Notably, with our lower dimension feature representation, we can still archive competitive results against these appearance-based methods.

Furthermore, our approach shows a high ability to model temporal features as shown in Tab~\ref{tab:temp-ablation}. When trained with sorted sequences and tested with shuffled sequences (row c), the performance drops profoundly. As a comparison, the same ablation study was conducted by GaitPart \cite{fan2020gaitpart}, with only a slight drop in performance from row b to c. These results further support our claim of bringing back real temporal features to gait recognition. Tab~\ref{tab:temp-ablation} also illustrates the spatial modeling abilities in row a. Despite the missing temporal and appearance information, the network is still able to learn appearance-invariant features of the person's underlying physic.

\begin{table}
 \caption{Spatio-temporal Study. Control Condition: shuffle/sort the input sequence at train/test phase. Results are rank-1 accuracies on CASIA-B averaged in percent.}
 \label{tab:temp-ablation}
\setlength{\tabcolsep}{5pt}
\fontsize{9}{11}\selectfont
 \centering
 
 \begin{tabularx}{.985\linewidth}{c | c c | c c c | c c c }
 \toprule
     & \multicolumn{2}{c|}{} & \multicolumn{3}{c|}{\textbf{GaitGraph}} & \multicolumn{3}{c}{GaitPart\cite{fan2020gaitpart}} \\
     & Train & Test & NM & BG & CL & NM & BG & CL\\
    \hline
    a & Shuffle & Sort & 47.3 & 36.9 & 26.9 & 95.6 & 89.9 & 71.5\\
    b & Sort & Sort & \textbf{87.7} & \textbf{74.8} & \textbf{66.3} & \textbf{96.2} & \textbf{91.5} & \textbf{78.7}\\
    c & Sort & Shuffle & 26.4 & 22.0 & 16.7 & 92.5 & 85.8 & 65.1\\
 \bottomrule
\end{tabularx}
\end{table}

\section{Conclusion}
In this paper, we present a novel approach to interpret gait as a sequence of skeleton graphs. Thus, GaitGraph is proposed, which uses a human pose estimator to extract the 2D skeleton pose, and extract the gait information considering the inherent graph structure of the skeleton.
Furthermore, experiments conducted on the well-known database CASIA-B \cite{yu2006framework} show \gls{sota} results in model-based gait recognition and competitive results against appearance-based methods in gait recognition. Our spatial-temporal ablations proves our claim to bring back true temporal gait features instead of mostly relying on the appearance. 

\vfill
\pagebreak

\bibliographystyle{IEEEbib}
\bibliography{refs}

\begin{thebibliography}{10}

\bibitem{song2019gaitnet}
Chunfeng Song, Yongzhen Huang, Yan Huang, Ning Jia, and Liang Wang,
\newblock ``{GaitNet}: An end-to-end network for gait based human
  identification,''
\newblock {\em Pattern Recognition}, vol. 96, pp. 106988, 2019.

\bibitem{chao2019gaitset}
Hanqing Chao, Yiwei He, Junping Zhang, and Jianfeng Feng,
\newblock ``{GaitSet}: Regarding gait as a set for cross-view gait
  recognition,''
\newblock in {\em AAAI}, 2019, vol.~33, pp. 8126--8133.

\bibitem{fan2020gaitpart}
Chao Fan, Yunjie Peng, Chunshui Cao, Xu~Liu, Saihui Hou, Jiannan Chi, Yongzhen
  Huang, Qing Li, and Zhiqiang He,
\newblock ``{GaitPart}: Temporal part-based model for gait recognition,''
\newblock in {\em CVPR}. IEEE, June 2020.

\bibitem{wang2003silhouette}
{Liang Wang}, {Tieniu Tan}, {Huazhong Ning}, and {Weiming Hu},
\newblock ``Silhouette analysis-based gait recognition for human
  identification,''
\newblock {\em PAMI}, vol. 25, no. 12, pp. 1505--1518, 2003.

\bibitem{wu2016comprehensive}
Zifeng Wu, Yongzhen Huang, Liang Wang, Xiaogang Wang, and Tieniu Tan,
\newblock ``A comprehensive study on cross-view gait based human identification
  with deep cnns,''
\newblock {\em PAMI}, vol. 39, no. 2, pp. 209--226, 2016.

\bibitem{liao2017pose}
Rijun Liao, Chunshui Cao, Edel~B Garcia, Shiqi Yu, and Yongzhen Huang,
\newblock ``Pose-based temporal-spatial network (ptsn) for gait recognition
  with carrying and clothing variations,''
\newblock in {\em Chinese Conference on Biometric Recognition}. Springer, 2017,
  pp. 474--483.

\bibitem{liao2020model}
Rijun Liao, Shiqi Yu, Weizhi An, and Yongzhen Huang,
\newblock ``A model-based gait recognition method with body pose and human
  prior knowledge,''
\newblock {\em Pattern Recognition}, vol. 98, pp. 107069, 2020.

\bibitem{cheng2020bottom}
Bowen Cheng, Bin Xiao, Jingdong Wang, Honghui Shi, Thomas~S. Huang, and Lei
  Zhang,
\newblock ``Higherhrnet: Scale-aware representation learning for bottom-up
  human pose estimation,''
\newblock in {\em CVPR}. IEEE, 2020.

\bibitem{yan2018spatial}
Sijie Yan, Yuanjun Xiong, and Dahua Lin,
\newblock ``Spatial temporal graph convolutional networks for skeleton-based
  action recognition,''
\newblock in {\em AAAI}, 2018, vol.~32.

\bibitem{song2020stronger}
Yi-Fan Song, Zhang Zhang, Caifeng Shan, and Liang Wang,
\newblock {\em Stronger, Faster and More Explainable: A Graph Convolutional
  Baseline for Skeleton-Based Action Recognition}, p. 1625–1633,
\newblock Association for Computing Machinery, New York, NY, USA, 2020.

\bibitem{bouchrika2007model}
Imed Bouchrika and Mark~S Nixon,
\newblock ``Model-based feature extraction for gait analysis and recognition,''
\newblock in {\em International Conference on Computer Vision / Computer
  Graphics Collaboration Techniques and Applications}. Springer, 2007, pp.
  150--160.

\bibitem{han2005individual}
Jinguang Han and Bir Bhanu,
\newblock ``Individual recognition using gait energy image,''
\newblock {\em PAMI}, vol. 28, no. 2, pp. 316--322, 2005.

\bibitem{wolf2016multi}
Thomas Wolf, Mohammadreza Babaee, and Gerhard Rigoll,
\newblock ``Multi-view gait recognition using 3d convolutional neural
  networks,''
\newblock in {\em ICIP}. IEEE, 2016, pp. 4165--4169.

\bibitem{sokolova2018pose}
Anna Sokolova and Anton Konushin,
\newblock ``Pose-based deep gait recognition,''
\newblock {\em IET Biometrics}, vol. 8, no. 2, pp. 134--143, 2018.

\bibitem{an2018improving}
Weizhi An, Rijun Liao, Shiqi Yu, Yongzhen Huang, and Pong~C Yuen,
\newblock ``Improving gait recognition with 3d pose estimation,''
\newblock in {\em Chinese Conference on Biometric Recognition}. Springer, 2018,
  pp. 137--147.

\bibitem{sun2019deep}
Ke~Sun, Bin Xiao, Dong Liu, and Jingdong Wang,
\newblock ``Deep high-resolution representation learning for human pose
  estimation,''
\newblock in {\em CVPR}. IEEE, June 2019.

\bibitem{lin2014microsoft}
Tsung-Yi Lin, Michael Maire, Serge Belongie, James Hays, Pietro Perona, Deva
  Ramanan, Piotr Doll{\'a}r, and C~Lawrence Zitnick,
\newblock ``Microsoft coco: Common objects in context,''
\newblock in {\em ECCV}. Springer, 2014, pp. 740--755.

\bibitem{khosla2020supervised}
Prannay Khosla, Piotr Teterwak, Chen Wang, Aaron Sarna, Yonglong Tian, Phillip
  Isola, Aaron Maschinot, Ce~Liu, and Dilip Krishnan,
\newblock ``Supervised contrastive learning,'' 2020.

\bibitem{yu2006framework}
Shiqi Yu, Daoliang Tan, and Tieniu Tan,
\newblock ``A framework for evaluating the effect of view angle, clothing and
  carrying condition on gait recognition,''
\newblock in {\em ICPR}. IEEE, 2006, vol.~4, pp. 441--444.

\bibitem{takemura2018multi}
Noriko Takemura, Yasushi Makihara, Daigo Muramatsu, Tomio Echigo, and Yasushi
  Yagi,
\newblock ``Multi-view large population gait dataset and its performance
  evaluation for cross-view gait recognition,''
\newblock {\em IPSJ Transactions on Computer Vision and Applications}, vol. 10,
  no. 1, pp. 4, Feb 2018.

\bibitem{zhang2019gait}
Ziyuan Zhang, Luan Tran, Xi~Yin, Yousef Atoum, Xiaoming Liu, Jian Wan, and
  Nanxin Wang,
\newblock ``Gait recognition via disentangled representation learning,''
\newblock in {\em CVPR}. IEEE, 2019, pp. 4710--4719.

\bibitem{smith2019super}
Leslie~N Smith and Nicholay Topin,
\newblock ``Super-convergence: Very fast training of neural networks using
  large learning rates,''
\newblock in {\em Artificial Intelligence and Machine Learning for Multi-Domain
  Operations Applications}. International Society for Optics and Photonics,
  2019, vol. 11006, p. 1100612.

\end{thebibliography}

\end{document}